\documentclass[letterpaper, 10 pt, conference]{ieeeconf}
\IEEEoverridecommandlockouts 
\usepackage{amsmath,amsfonts,amssymb}
\usepackage[protrusion=true,spacing=true]{microtype}
\usepackage[bookmarks=true]{hyperref}
\usepackage[noend]{algorithmic}
\usepackage{algorithm}
\usepackage{graphicx}
\usepackage{setspace}
\usepackage[font=small]{caption}
\usepackage{subcaption}
\usepackage{tabularx}
\usepackage{multirow}

\newcommand{\Ve}{\mathcal{I}}
\newcommand{\St}{\mathcal{S}}
\newcommand{\st}{s}
\newcommand{\tr}{\hat{\st}}
\newcommand{\Ob}{\mathcal{O}}
\newcommand{\ob}{o}
\newcommand{\Ac}{\mathcal{A}}
\newcommand{\ac}{a}
\newcommand{\go}{G}
\newcommand{\Go}{\mathcal{G}}
\newcommand{\eg}{\varepsilon}
\newcommand{\ma}{\omega}
\newcommand{\man}[1]{\textit{#1}}

\graphicspath{{figures/}}

\begin{document}

    \title{Interpretable Goal-based Prediction and Planning for Autonomous Driving}

    \author{\authorblockN{Stefano V. Albrecht\authorrefmark{1}\authorrefmark{2},
        Cillian Brewitt\authorrefmark{1}\authorrefmark{2},
        John Wilhelm\authorrefmark{1}\authorrefmark{2},
        Balint Gyevnar\authorrefmark{1}\authorrefmark{2}, \\
        Francisco Eiras\authorrefmark{1}\authorrefmark{3},
        Mihai Dobre\authorrefmark{1},
        Subramanian Ramamoorthy\authorrefmark{1}\authorrefmark{2}}
        \authorblockA{\authorrefmark{1}Five\,AI Ltd., UK, \{firstname.lastname\}@five.ai}
        \authorblockA{\authorrefmark{2}School of Informatics, University of Edinburgh, UK}
        \authorblockA{\authorrefmark{3}Department of Engineering Science, University of Oxford, UK}
        \thanks{S.A. is supported by a Royal Society Industry Fellowship. C.B., J.W., B.G. were interns at Five\,AI with partial financial support from the Royal Society and UKRI. {\bf IGP2 code:} \url{https://github.com/uoe-agents/IGP2}}
        \vspace{-1.0em}
    }
    
    \maketitle

    \begin{abstract}
        We propose an integrated prediction and planning system for autonomous driving which uses rational inverse planning to recognise the goals of other vehicles. Goal recognition informs a Monte Carlo Tree Search (MCTS) algorithm to plan optimal maneuvers for the ego vehicle. Inverse planning and MCTS utilise a shared set of defined maneuvers and macro actions to construct plans which are explainable by means of \emph{rationality} principles. Evaluation in simulations of urban driving scenarios demonstrate the system's ability to robustly recognise the goals of other vehicles, enabling our vehicle to exploit non-trivial opportunities to significantly reduce driving times. In each scenario, we extract intuitive explanations for the predictions which justify the system's decisions.
    \end{abstract}

    \section{Introduction}
    \label{sec:intro}

The ability to predict the intentions and driving trajectories of other vehicles is a key problem for autonomous driving \cite{sar2018}. This problem is significantly complicated by the need to make fast and accurate predictions based on limited observation data which originate from coupled multi-agent interactions.

To make prediction tractable in such conditions, a standard approach in autonomous driving research is to assume that vehicles use one of a finite number of distinct high-level maneuvers, such as lane-follow, lane-change, turn, stop, etc. \cite{dong2018,hubmann2018,zhou2018,mpdm2017,hubmann2017,song2016}. A classifier of some type is used to detect a vehicle's current executed maneuver based on its observed driving trajectory. The limitation in such methods is that they only detect the \emph{current} maneuver of other vehicles, hence planners using such predictions are effectively limited to the timescales of the detected maneuvers.
An alternative approach is to specify a finite set of possible \emph{goals} for each other vehicle (such as road exit points) and to plan a full trajectory to each goal from the vehicle's observed local state \cite{ziebart2009planning,hardy2013,bandy2013}. While this approach can generate longer-term predictions, a limitation is that the generated trajectories must be matched relatively closely by a vehicle in order to yield high-confidence predictions of the vehicle's goals.

Recent methods based on deep learning have shown promising results for trajectory prediction in autonomous driving \cite{tnt2020,precog2019,multipath2019,xu2019,intentnet2018,lee2017,wulfmeier2016}.
Prediction models are trained on large datasets that are becoming available through data gathering campaigns involving sensorised vehicles (e.g. video, lidar, radar).
Reliable prediction over several second horizons remains a hard problem, in part due to the difficulties in capturing the coupled evolution of traffic.
In our view, one of the most significant limitations of this class of methods (though see recent progress \cite{sadat2020perceive}) is the difficulty in extracting interpretable predictions in a form that is amenable to efficient integration with planning methods that effectively represent multi-dimensional and hierarchical task objectives.

Our starting point is that in order to predict the future maneuvers of a vehicle, we must reason about \emph{why} -- that is, to what end -- the vehicle performed its past maneuvers, which will yield clues as to its intended goal \cite{as2018aij}. Knowing the goals of other vehicles enables prediction of their future maneuvers and trajectories, which facilitates planning over extended timescales.
We show in our work (illustrated in Figure~\ref{fig:scenarios}) how such reasoning can help to address the problem of overly-conservative autonomous driving \cite{wired2020}.
Further, to the extent that our predictions are structured around the interpretation of observed trajectories in terms of high-level maneuvers, the goal recognition process lends itself to \emph{intuitive interpretation} for the purposes of system analysis and debugging, at a level of detail suggested in Figure~\ref{fig:scenarios}. As we develop towards making our autonomous systems more trustworthy \cite{fivesafety2019}, these notions of interpretation and the ability to justify (explain) the system's decisions are key \cite{gadd2020sense}.

To this end, we propose \emph{Interpretable Goal-based Prediction and Planning} (IGP2) which leverages the computational advantages of using a finite space of maneuvers, but extends the approach to planning and prediction of \emph{sequences} (i.e., plans) of maneuvers. We achieve this via a novel integration of rational inverse planning \cite{rg2010,bst2009} to recognise the goals of other vehicles, with Monte Carlo Tree Search (MCTS)~\cite{bpwl2012} to plan optimal maneuvers for the ego vehicle. Inverse planning and MCTS utilise a shared set of defined maneuvers to construct plans which are explainable by means of \emph{rationality} principles, i.e. plans are optimal with respect to given metrics.
We evaluate IGP2 in simulations of diverse urban driving scenarios, showing that (1) the system robustly recognises the goals of other vehicles, even if significant parts of a vehicle's trajectory are occluded, (2) goal recognition enables our vehicle to exploit opportunities to improve driving efficiency as measured by driving time compared to other prediction baselines, and (3) we are able to extract intuitive explanations for the predictions to justify the system's decisions.

{\bf In summary, our contributions are:}
\begin{itemize}
	\item A method for goal recognition and multi-modal trajectory prediction via rational inverse planning.
	\item Integration of goal recognition with MCTS planning to generate optimised plans for the ego vehicle.
	\item Evaluation in simulated urban driving scenarios showing accurate goal recognition, improved driving efficiency, and ability to interpret the predictions and ego plans.
\end{itemize}

    \section{Preliminaries and Problem Definition}
    \label{sec:problem}

Let $\Ve$ be the set of vehicles in the local neighbourhood of the ego vehicle (including itself). At time $t$, each vehicle $i \in \Ve$ is in a local state $\st_t^i \in \St^i$, receives a local observation $\ob_t^i \in \Ob^i$, and can choose an action $\ac_t^i \in \Ac^i$. We write $\st_t \in \St = \times_i \St^i$ for the joint state and $\st_{a:b}$ for the tuple $(\st_a,...,\st_b)$, and similarly for $\ob_t \in \Ob, \ac_t \in \Ac$. Observations depend on the joint state via $p(\ob_t^i | \st_t)$, and actions depend on the observations via $p(\ac_t^i | \ob_{1:t}^i)$. In our system, a local state contains a vehicle's pose, velocity, and acceleration (we use the terms velocity and speed interchangeably); an observation contains the poses and velocities of nearby vehicles; and an action controls the vehicle's steering and acceleration. The probability of a sequence of joint states $\st_{1:n}$ is given by
\begin{equation}
    p(\st_{1:n}) = \prod_{t=1}^{n-1} \int_\Ob \int_\Ac p(\ob_t | \st_t) p(\ac_t | \ob_{1:t}) p(\st_{t+1} | \st_t, \ac_t) \, d \ob_t \, d \ac_t
\end{equation}
where $p(\st_{t+1} | \st_t, \ac_t)$ defines the joint vehicle dynamics, and we assume independent local observations and actions, $p(\ob_t | \st_t) = \prod_i p(\ob_t^i | \st_t)$ and $p(\ac_t | \ob_{1:t}) = \prod_i p(\ac_t^i | \ob_{1:t}^i)$. Vehicles react to other vehicles via their observations $\ob_{1:n}^i$.

We define the planning problem as finding an optimal policy $\pi^*$ which selects the actions for the ego vehicle, $\eg$, to achieve a specified goal, $\go^\eg$, while optimising the driving trajectory via a defined reward function. Here, a policy is a function $\pi : ( \Ob^\eg )^* \mapsto \Ac^\eg$ which maps an observation sequence $o^\eg_{1:n}$ to an action $\ac_t^\eg$. (State filtering \cite{albrecht2016causality} is outside the scope of this work.) A goal can be any subset of local states, $\go^\eg \subset \mathcal{S}^\eg$. In this paper, we focus on goals that specify target locations and ``stopping goals'' which specify a target location and zero velocity. Formally, define
\begin{equation} \label{eq:omega}
    \Omega_n = \left\{ \st_{1:n} \,\big|\, \st_n^\eg \in \go^\eg \land \forall m < n : \st_m^\eg \not\in \go^\eg \right\}
\end{equation}
where $\st_n^\eg \in \go^\eg$ means that $\st_n^\eg$ satisfies $\go^\eg$. The second condition in \eqref{eq:omega} ensures that $\sum_{n=1}^{\infty} \int_{\Omega_n} p(\st_{1:n}) d \st_{1:n} \leq 1$ for any policy $\pi$, which is needed for soundness of the sum in \eqref{eq:policyopt}. The problem is to find $\pi^*$ such that
\begin{equation}\label{eq:policyopt}
    \pi^* \in \arg\max_{\pi} \sum_{n=1}^{\infty} \int_{\Omega_n} p(\st_{1:n}) R^\eg(\st_{1:n}) \, d \st_{1:n}
\end{equation}
where $R^i(\st_{1:n})$ is vehicle~$i$'s reward for $\st_{1:n}$. We define $R^i$ as a weighted sum of reward elements based on trajectory execution time, longitudinal and lateral jerk, path curvature, and safety distance to leading vehicle.

    \section{IGP2: Interpretable Goal-based Prediction and Planning}
    \label{sec:method}

\begin{figure}[t]
    \centering
    \includegraphics[width=0.48\textwidth]{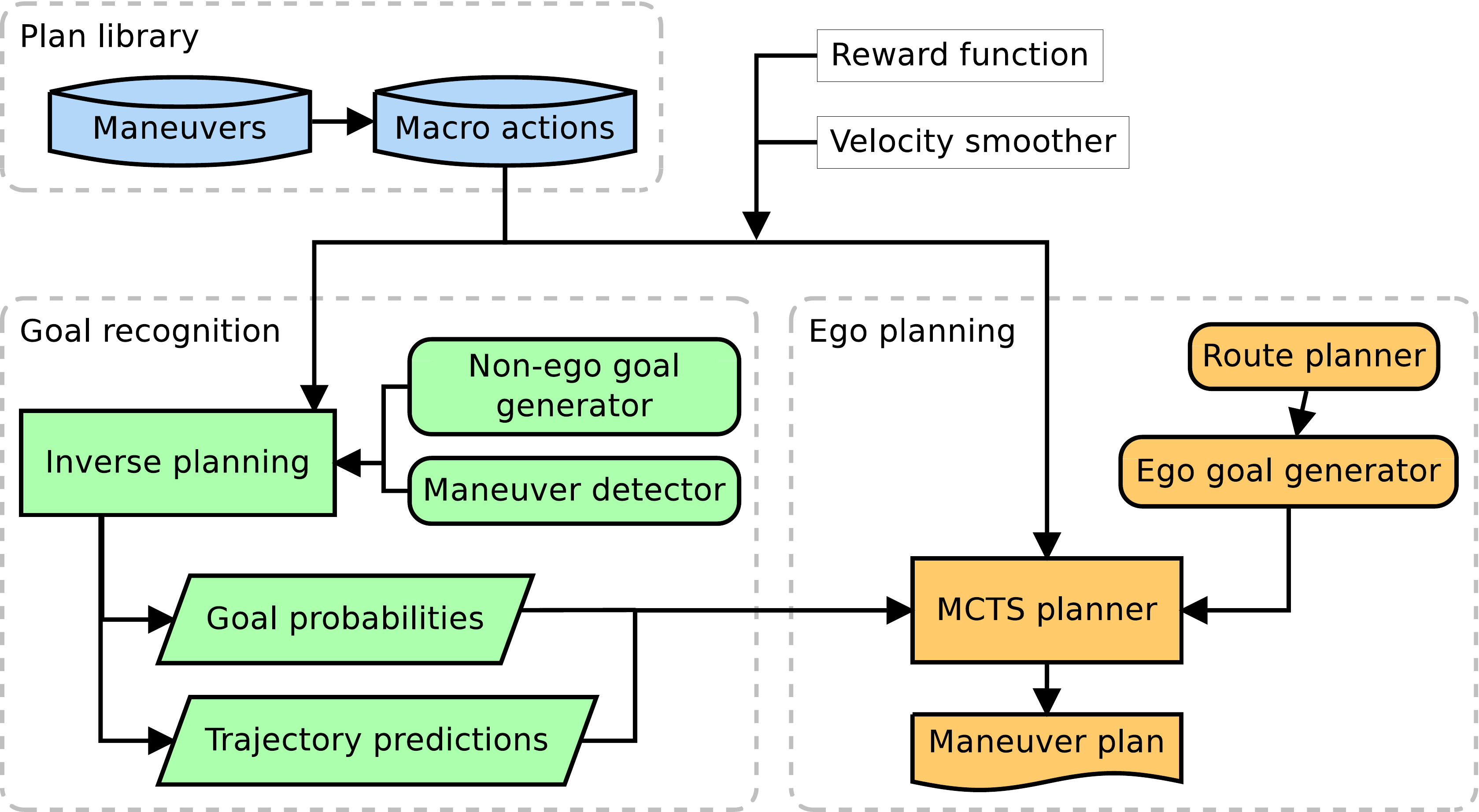}
    \caption{IGP2 system overview.}
    \label{fig:overview}
    \vspace{-1.0em}
\end{figure}

Our general approach relies on two assumptions: (1) each vehicle seeks to reach some (unknown) goal from a set of possible goals, and (2) each vehicle follows a plan generated from a finite library of defined maneuvers.

Figure~\ref{fig:overview} provides an overview of the components in our proposed IGP2 system. At a high level, IGP2 approximates the optimal ego policy $\pi^*$ as follows: For each non-ego vehicle, generate its possible goals and inversely plan for that vehicle to each goal. The resulting goal probabilities and predicted trajectories for each non-ego vehicle inform the simulation process of a Monte Carlo Tree Search (MCTS) algorithm, to generate an optimal maneuver plan for the ego vehicle toward its current goal. In order to keep the required search depth in inverse planning and MCTS shallow (and thus efficient), both plan over a shared set of macro actions which flexibly concatenate maneuvers using context information. We detail these components in the following sections.

\begin{table*}[t]
    \centering
    \fontsize{8.6}{11}\selectfont
    \begin{tabular}{@{} l @{\hspace{6pt}} l @{\hspace{6pt}} l @{}}
        \hline
        {\bf Macro action:} & {\bf Additional applicability condition:} & {\bf Maneuver sequence (maneuver parameters in brackets):} \\
        \hline
        \man{Continue}                &   ---                                                                        & \man{lane-follow} (end of visible lane) \\
        \man{Continue next exit}   & Must be in roundabout and not in outer-lane                               & \man{lane-follow} (next exit point) \\
        \man{Change left/right}       & There is a lane to the left/right                                         & \man{lane-follow} (until target lane clear), \man{lane-change-left/right} \\
        \man{Exit left/right}         & Exit point on same lane ahead of car and in correct direction   & \man{lane-follow} (exit point), \man{give-way} (relevant lanes), \man{turn-left/right} \\
        \man{Stop}                    & There is a stopping goal ahead of the car on the current lane             & \man{lane-follow} (close to stopping point), \man{stop} \\
        \hline
    \end{tabular}
    \caption{Macro actions used in our system. Each macro action concatenates one or more maneuvers and automatically sets their parameters.}
    \label{tab:macros}
    \vspace{-0.8em}
\end{table*}

        \subsection{Maneuvers}
        \label{sec:maneuvers}

We assume that at any time, each vehicle is executing one of the following maneuvers: \man{lane-follow, lane-change-left/right, turn-left/right, give-way, stop.} Each maneuver $\ma$ specifies applicability and termination conditions.
For example, lane-change-left is only applicable if there is a lane in same driving direction to the left of the vehicle, and terminates once the vehicle has reached the new lane and its orientation is aligned with the lane. Some maneuvers have free parameters, e.g. \man{follow-lane} has a parameter to specify when to terminate.

If applicable, a maneuver specifies a local trajectory $\tr_{1:n}^i$ to be followed by the vehicle, which includes a reference path in the global coordinate frame and target velocities along the path. For convenience in exposition, we assume that $\tr^i$ uses the same representation and indexing as $\st^i$, but in general this does not have to be the case (for example, $\tr$ may be indexed by longitudinal position rather than time, which can be interpolated to time indices). In our system, the reference path is generated via a Bezier spline function fitted to a set of points extracted from the road topology, and target velocities are set using domain heuristics similar to \cite{gamez2017}.

        \subsection{Macro Actions}
        \label{sec:macro}

Macro actions specify common sequences of maneuvers and automatically set the free parameters (if any) in maneuvers based on context information such as road layout. Table~\ref{tab:macros} specifies the macro actions used in our system. The applicability condition of a macro action is given by the applicability condition of the first maneuver in the macro action as well as optional additional conditions. The termination condition of a macro action is given by the termination condition of the last maneuver in the macro action.

        \subsection{Velocity Smoothing}
        \label{sec:smoother}

To obtain a feasible trajectory across maneuvers for vehicle $i$, we define a velocity smoothing operation which optimises the target velocities in a given trajectory $\tr_{1:n}^i$. Let $\hat{x}_t$ be the longitudinal position on the reference path at $\tr_t^i$ and $\hat{v}_t$ its target velocity, for $1 \leq t \leq n$. We define $\kappa: x \rightarrow v$ as the piecewise linear interpolation of target velocities between points $\hat{x}_t$. Given the time elapsed between two time steps, $\Delta t$; the maximum velocity and acceleration, $v_{max}$/$a_{max}$; and setting $x_1 = \hat{x}_1, v_1 = \hat{v}_1$, we define velocity smoothing as
\vspace{-5pt}
\begin{equation} \label{eq:smooth}
\begin{aligned}
& &\min_{x_{2:n}, v_{2:n}} & & & \sum_{t=1}^n ||v_t - \kappa(x_t)||_2 +  \lambda\sum_{t=1}^{n-1} \left|\left|v_{t+1} - v_t\right|\right|_2 \\
& & & & & \text{s.t. } x_{t+1} = x_t + v_t \Delta t \\
& & & & & 0 < v_t < v_{\max}, \ v_t \leq \kappa(x_t) \\
& & & & & | v_{t+1} - v_t| < a_{\max} \Delta t
\end{aligned}
\end{equation}
where $\lambda > 0$ is the weight given to the acceleration part of the optimisation objective. Eq.~\eqref{eq:smooth} is a nonlinear non-convex optimisation problem which can be solved, e.g., using a primal-dual interior point method (we use IPOPT \cite{ipopt}). From the solution of the problem, $(x_{2:n}, v_{2:n})$, we interpolate to obtain the achievable velocities at the original points $\hat{x}_t$.

        \subsection{Goal Recognition}
        \label{sec:goalrec}

We assume that each non-ego vehicle $i$ seeks to reach one of a finite number of possible goals $\go^i \in \Go^i$, using plans constructed from our defined macro actions. We use the framework of rational inverse planning \cite{rg2010,bst2009} to compute a Bayesian posterior distribution over $i$'s goals at time $t$
\begin{equation}
    p(\go^i | \st_{1:t}) \propto L(\st_{1:t} | \go^i) p(\go^i)
\end{equation}
where $L(\st_{1:t} | \go^i)$ is the likelihood of $i$'s observed trajectory assuming its goal is $\go^i$, and $p(\go^i)$ specifies the prior probability of $\go^i$.

The likelihood is a function of the reward difference between two plans: the reward $\hat{r}$ of the optimal trajectory from $i$'s initial observed state $\st_1^i$ to goal $\go^i$ after velocity smoothing, and the reward $\bar{r}$ of the trajectory which follows the observed trajectory until time $t$ and then continues optimally to goal $\go^i$, with smoothing applied only to the trajectory after~$t$. The likelihood is defined as
\begin{equation} \label{eq:lik}
    L(\st_{1:t} | \go^i) = \exp( \beta (\bar{r} - \hat{r}) )
\end{equation}
where $\beta$ is a scaling parameter (we use $\beta = 1$). This likelihood definition assumes that vehicles drive approximately \emph{rationally} (i.e., optimally) to achieve their goals while allowing for some deviation. If a goal is infeasible, we set its probability to zero.

Algorithm~\ref{alg:goalrec} shows the pseudo code for our goal recognition algorithm, with further details in below subsections.

            \subsubsection{\bf Goal Generation}

A heuristic function is used to generate a set of possible goals $\Go^i$ for vehicle $i$ based on its location and context information such as road layout. In our system, we include goals for the visible end of the current road and connecting roads (bounded by the ego vehicle's view region). In addition to such static goals, it is also possible to add dynamic goals which depend on current traffic. For example, in the dense merging scenario shown in Figure~\ref{fig:s4}, stopping goals are dynamically added to model a vehicle's intention to allow the ego vehicle to merge in front.

            \subsubsection{\bf Maneuver Detection} \label{sec:mandetect}

Maneuver detection is used to detect the current executed maneuver of a vehicle (at time~$t$), allowing inverse planning to complete the maneuver before planning onward. We assume a module which computes probabilities over current maneuvers, $p(\ma^i)$, for each vehicle~$i$. One option is Bayesian changepoint detection (e.g. \cite{champ2015}). The details of maneuver detection are outside the scope of our paper and in our experiments we use a simulated detector (cf. Sec~\ref{sec:exp-algos}). As different current maneuvers may hint at different goals, we perform inverse planning for each possible current maneuver for which $p(\ma^i) > 0$. Thus, each current maneuver produces its associated posterior probabilities over goals, denoted by $p(\go^i \,|\, \st_{1:t}, \omega^i)$.

            \subsubsection{\bf Inverse Planning}
            \label{sec:astar}

Inverse planning is done using A* search \cite{hnr1968} over macro actions. A* starts after completing the current maneuver $\omega^i$ which produces the initial trajectory $\tr_{1:\tau}$. Each search node $q$ corresponds to a state $\st \in \St$, with initial node at state $\tr_\tau$, and macro actions are filtered by their applicability conditions applied to $\st$. A* chooses the next macro action leading to a node $q'$ which has lowest estimated total cost\footnote{Here we use the term ``cost'' in keeping with standard A* terminology and to differentiate from the reward function defined in Sec.~\ref{sec:problem}.} to goal $\go^i$, given by $f(q') = l(q') + h(q')$. The cost $l(q')$ to reach node $q'$ is given by the driving time from $i$'s location in the initial search node to its location in $q'$, following the trajectories returned by the macro actions leading to $q'$.
A* uses the assumption that all other vehicles not planned for use a constant-velocity lane-following model after their observed trajectories. We do not check for collisions during inverse planning. The cost heuristic $h(q')$ to estimate remaining cost from $q'$ to goal $\go^i$ is given by the driving time from $i$'s location in $q'$ to goal via straight line at speed limit. This definition of $h(q')$ is admissible as per A* theory, which ensures that the search returns an optimal plan. After the optimal plan is found, we extract the complete trajectory $\tr_{1:n}^i$ from the maneuvers in the plan and initial segment $\tr_{1:\tau}$.

            \subsubsection{\bf Trajectory Prediction}

Our system predicts multiple plausible trajectories for a given vehicle and goal. This is required because there are situations in which different trajectories may be (near-)optimal but may lead to different predictions which could require different behaviour on the part of the ego vehicle. We run A* search for a fixed amount of time and let it compute a set of plans with associated rewards (up to some fixed number of plans). Any time A* search finds a node that reaches the goal, the corresponding plan is added to the set of plans. Given a set of smoothed trajectories $\{ \tr_{1:n}^{i,k} | \omega^i, \go^i \}_{k=1..K}$ to goal $\go^i$ with initial maneuver $\omega^i$ and associated reward $r_k = R^i(\tr_{1:n}^{i,k})$, we compute a distribution over the trajectories via a Boltzmann distribution
\begin{equation}\label{eq:boltz}
    p(\tr_{1:n}^{i,k}) \propto \exp( \gamma \, r_k )
\end{equation}
where $\gamma$ is a scaling parameter (we use $\gamma = 1$). Similar to Eq.~\eqref{eq:lik}, Eq.~\eqref{eq:boltz} encodes the assumption that trajectories which are closer to optimal are more likely.

\begin{algorithm}[t]
    \fontsize{9}{11.0}\selectfont
    
	\textbf{Input:} vehicle $i$, current maneuver $\omega^i$, observations  $\st_{1:t}$ \\
	\textbf{Returns:} goal probabilities $p(\go^i | \st_{1:t}, \omega^i)$
	
	\begin{algorithmic}[1]
		\algsetup{linenodelimiter=: }
		\STATE Generate possible goals $\go^i \in \Go^i$ from state $\st_t^i$
		\STATE Set prior probabilities $p(\go^i)$ (e.g. uniform)
		\FORALL{$\go^i \in \Go^i$}
		    \STATE $\tr_{1:n}^i \gets$ \textsc{A*}($\omega^i$) from $\tr_1^i = \st_1^i$ to $\go^i$
		    \STATE Apply velocity smoothing to $\tr_{1:n}^i$
		    \STATE $\hat{r} \gets$ reward $R^i(\tr_{1:n}^i)$
		    \STATE $\bar{\st}_{1:m}^i \gets$ \textsc{A*}($\omega^i$) from $\bar{\st}_t^i$ to $\go^i$, with $\bar{\st}_{1:t}^i = \st_{1:t}^i$
		    \STATE Apply velocity smoothing to $\bar{\st}_{t+1:m}^i$
		    \STATE $\bar{r} \gets$ reward $R^i(\bar{\st}_{1:m}^i)$
		    \STATE $L(\st_{1:t} | \go^i,\omega^i) \gets \exp( \beta (\bar{r} - \hat{r}) )$
		\ENDFOR
		\STATE Return $p(\go^i | \st_{1:t},\omega^i) \propto L(\st_{1:t} | \go^i,\omega^i) \, p(\go^i)$
	\end{algorithmic}
	\caption{Goal recognition algorithm}
	\label{alg:goalrec}
\end{algorithm}

\begin{algorithm}[t]
    \fontsize{9}{11.0}\selectfont
    
	\textbf{Returns:} optimal maneuver for ego vehicle $\eg$ in state $s_t$ \\
	Perform $K$ simulations:
	\begin{algorithmic}[1]
		\algsetup{linenodelimiter=: }
		\STATE Search node $q.\st \gets \st_t$ ($root$ node)
		\STATE Search depth $d \gets 0$
		\FORALL{$i \in \Ve \setminus \{ \eg \}$}
		    \STATE Sample current maneuver $\omega^i \sim p(\omega^i)$
		    \STATE Sample goal $\go^i \sim p(\go^i \,|\, \st_{1:t}, \omega^i)$
		    \STATE Sample trajectory $\tr_{1:n}^i \in \{ \tr_{1:n}^{i,k} \,|\, \omega^i, \go^i \}$ with $p(\tr_{1:n}^{i,k})$
		\ENDFOR
		\WHILE{$d < d_{max}$}
		    \STATE Select macro action $\mu$ for $\eg$ applicable in $q.\st$ \label{alg:mcts-1}
		    \STATE $\tr_{\tau:\iota} \gets$ Simulate $\mu$  until it terminates, with non-ego vehicles following their sampled trajectories $\tr_{1:n}^i$
		    \STATE $r \gets \emptyset$
		    \IF{ego vehicle collides during $\tr_{\tau:\iota}$}
		        \STATE $r \gets r_{coll}$
		    \ELSIF{$\tr_\iota^\eg$ achieves ego goal $\go^\eg$}
		        \STATE $r \gets R^\eg(\tr_{t:n})$
		    \ELSIF{$d = d_{max}-1$}
		        \STATE $r \gets r_{term}$
		    \ENDIF
		    \IF{$r \neq \emptyset$}
		        \STATE Use \eqref{eq:backprop} to backprop $r$ along search branches $(q,\mu,q')$ that generated the simulation
		        \STATE Start next simulation
		    \ENDIF
		    \STATE $q'.s = \tr_\iota$; $q \gets q'$; $d \gets d + 1$
		\ENDWHILE
	\end{algorithmic}
	Return maneuver for $\eg$ in $\st_t$, $\mu \in \arg\max_\mu Q(root,\mu)$
	\caption{Monte Carlo Tree Search algorithm}
	\label{alg:mcts}
\end{algorithm}

    \begin{figure*}[t]
        \centering
        \begin{subfigure}[b]{0.24\textwidth}
            \centering
            \includegraphics[width=\textwidth]{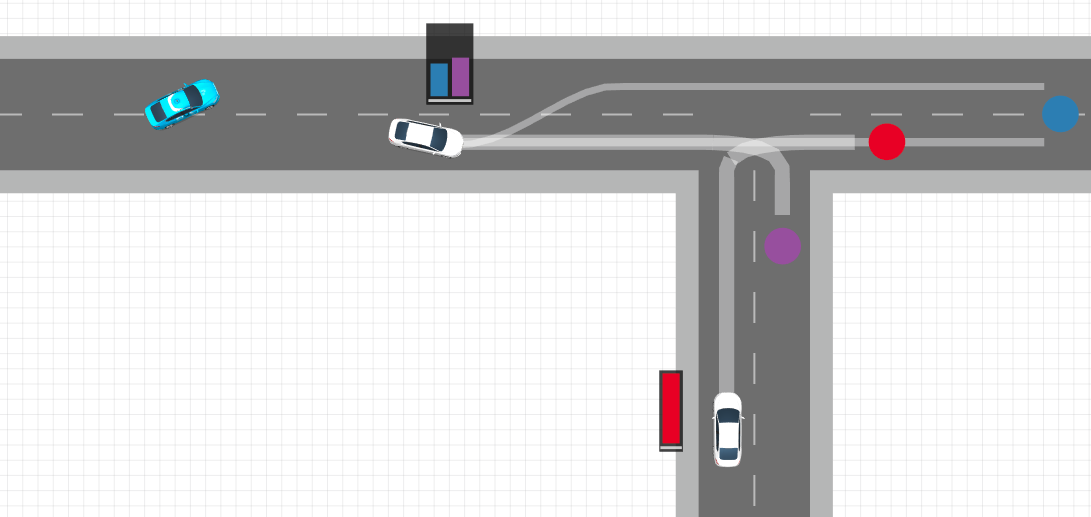}
             \caption{Scenario 1 (S1)}
            \label{fig:s1}
        \end{subfigure}
        \hfill
        \begin{subfigure}[b]{0.26\textwidth}
            \centering
            \includegraphics[width=\textwidth]{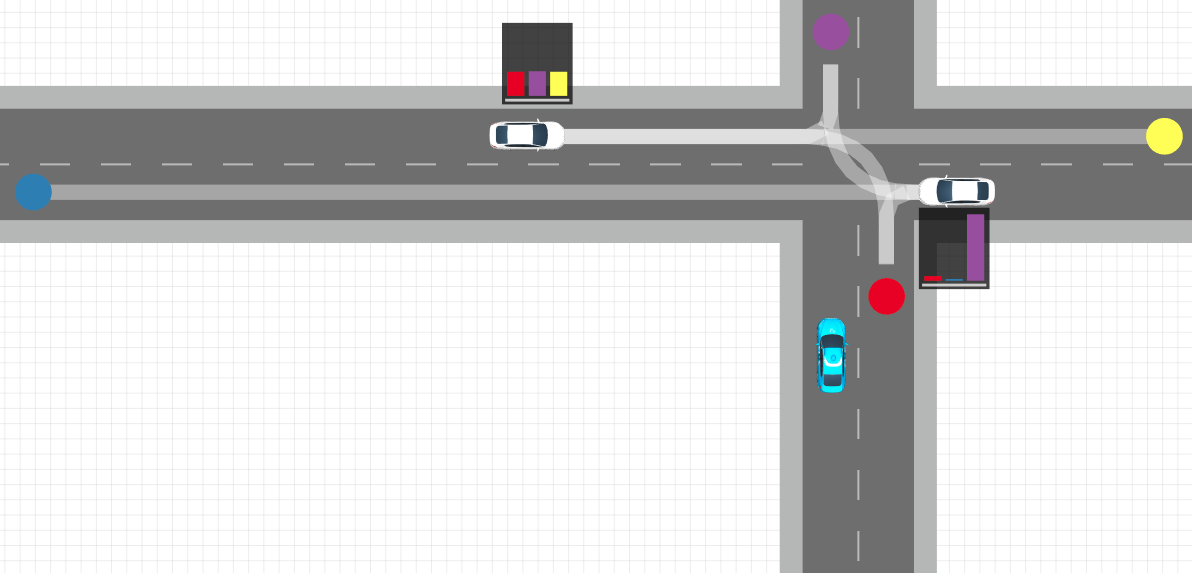}
             \caption{Scenario 2 (S2)}
            \label{fig:s2}
        \end{subfigure}
        \hfill
        \begin{subfigure}[b]{0.185\textwidth}
            \centering
            \includegraphics[width=\textwidth]{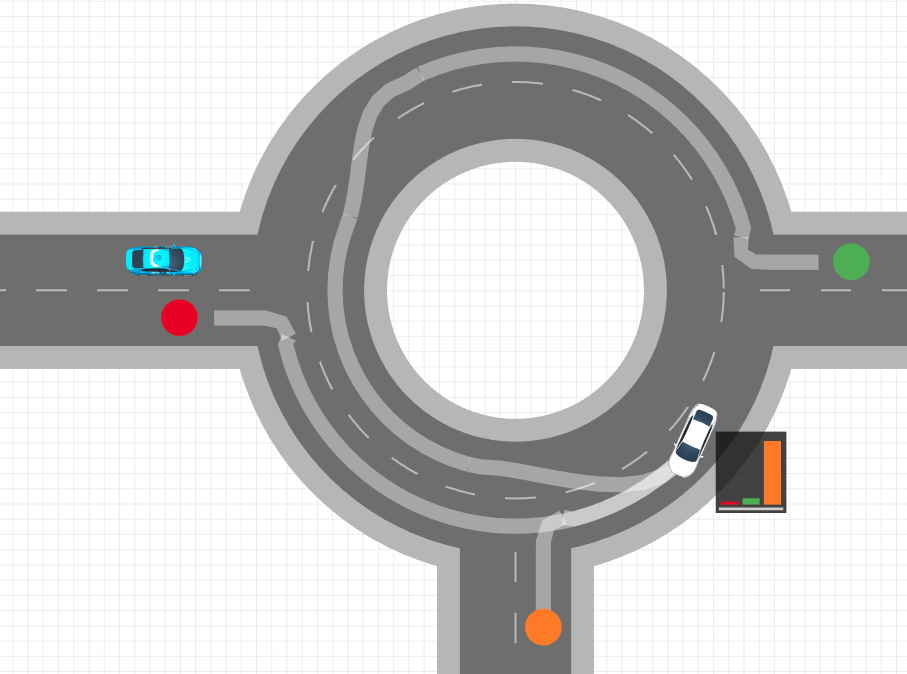}
             \caption{Scenario 3 (S3)}
            \label{fig:s3}
        \end{subfigure}
        \hfill
        \begin{subfigure}[b]{0.21\textwidth}
            \centering
            \includegraphics[width=\textwidth]{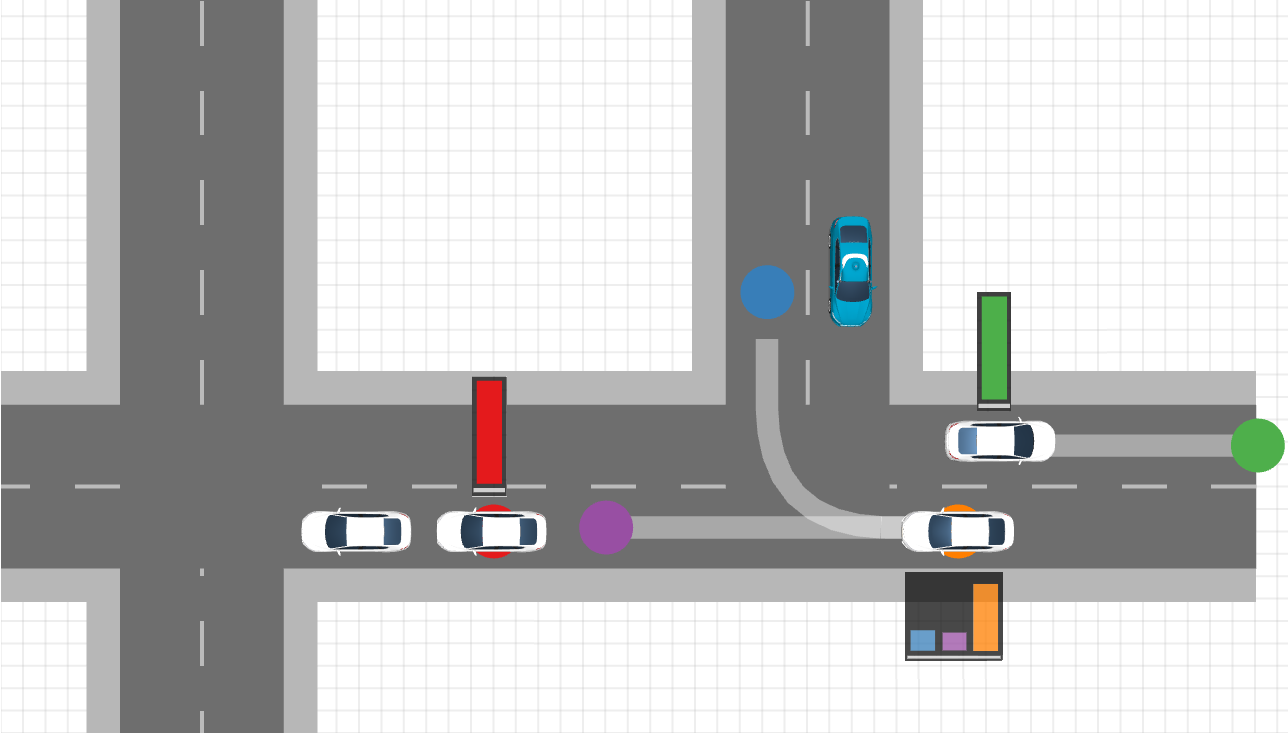}
             \caption{Scenario 4 (S4)}
            \label{fig:s4}
        \end{subfigure}
        \hfill
        \caption{{\bf IGP2 in 4 test scenarios.} Ego vehicle shown in blue. Bar plots show goal probabilities for non-ego vehicles. For each goal, up to two of the most probable predicted trajectories to goal are shown with thickness proportional to probability. (a) \textbf{S1}: Ego's goal is blue goal. Vehicle $V_1$ is on the ego's road, $V_1$ changes from left to right lane, biasing the ego prediction towards the belief that $V_1$ will exit, since a lane change would be irrational if $V_1$'s goal was to go east. As exiting will require a significant slowdown, the ego decides to switch lanes to avoid being slowed down too. (b) \textbf{S2}: Ego's goal is blue goal. Vehicle $V_1$ is approaching the junction from the east and vehicle $V_2$ from the west. As $V_1$ approaches the junction, slows down and waits to take a turn, the ego's belief that $V_1$ will turn right increases significantly, since it would be irrational to stop if the goal was to turn left or go straight. Since the ego recognised $V_1$'s goal is to go north, it predicts that $V_1$ will wait until $V_2$ has passed, giving the ego an opportunity to enter the road. (c) \textbf{S3}: Ego's goal is green goal. As $V_1$ changes from the inside to the outside lane of the roundabout and decreases its speed, it significantly biases the ego prediction towards the belief that $V_1$ will take the south exit since that is the rational course of action for that goal. This encourages the ego to enter the roundabout while $V_1$ is still in roundabout. (d) \textbf{S4}: Ego's goal is purple goal. With two vehicles stopped at the junction at a traffic light, vehicle $V_1$ is approaching them from behind, and vehicle $V_2$ is crossing in the opposite direction. When $V_1$ reaches zero velocity, the goal generation function adds a stopping goal (orange) for $V_1$ in its current position, shifting the goal distribution towards it since stopping is not rational for the north/west goals. The interpretation is that $V_1$ wants the ego to merge in front of $V_1$, which the ego then does.}
        \label{fig:scenarios}
        \vspace{-0.5em}
    \end{figure*}

        \subsection{Ego Vehicle Planning}
        \label{sec:mcts}

To compute an optimal plan for the ego vehicle, we use the goal probabilities and predicted trajectories to inform a Monte Carlo Tree Search (MCTS) algorithm \cite{bpwl2012} (see Algorithm~\ref{alg:mcts}).

The algorithm performs a number of closed-loop simulations $\tr_{t:n}$, starting in the current state $\tr_t = \st_t$ down to some fixed search depth or until a goal state is reached. At the start of each simulation, for each non-ego vehicle, we first sample a current maneuver, then goal, and then trajectory for the vehicle using the associated probabilities (cf. Section~\ref{sec:goalrec}).
Each node $q$ in the search tree corresponds to a state $\st \in \St$ and macro actions are filtered by their applicability conditions applied to $\st$. After selecting a macro action $\mu$ using some exploration technique (we use UCB1 \cite{ucb2002}), the state in the current search node is forward-simulated based on the trajectory generated by the macro action $\mu$ and the sampled trajectories of non-ego vehicles, resulting in a partial trajectory $\tr_{\tau:\iota}$ and new search node~$q'$ with state $\tr_{\iota}$.
Forward-simulation of trajectories uses a combination of proportional control and adaptive cruise control (based on IDM \cite{idm}) to control a vehicle's acceleration and steering. Termination conditions of maneuvers are monitored in each time step based on the vehicle's observations.
Collision checking is performed on $\tr_{\tau:\iota}$ to check whether the ego vehicle collided, in which case we set the reward to $r \gets r_{coll}$ which is back-propagated using \eqref{eq:backprop}, where $r_{coll}$ is a method parameter. Otherwise, if the new state $\tr_{\iota}$ achieves the ego goal $\go^\eg$, we compute the reward for back-propagation as $r \gets R^\eg(\tr_{t:n})$. If the search reached its maximum depth $d_{max}$ without colliding or achieving the goal, we set $r \gets r_{term}$ which can be a constant or based on heuristic reward estimates similar to A* search.

The reward $r$ is back-propagated through search branches $(q,\mu,q')$ that generated the simulation, using a 1-step off-policy update function (similar to Q-learning \cite{watkins1992})
\begin{equation} \label{eq:backprop}
    Q(q,\mu) \gets Q(q,\mu) + \left\{ \hspace{-3pt} \begin{array}{l}
        \delta^{-1} [ r - Q(q,\mu) ] \ \text{if $q$ leaf node, else} \\[3pt]
        \delta^{-1} [ \max_{\mu'} Q(q',\mu') - Q(q,\mu) ]
    \end{array} \right.
\end{equation}
where $\delta$ is the number of times that macro action $\mu$ has been selected in $q$. After the simulations are completed, the algorithm selects the best macro action for execution in $s_t$ from the root node, $\arg\max_\mu Q(root,\mu)$.

    \section{Evaluation}
    \label{sec:eval}

We evaluate IGP2 in simulations of diverse urban driving scenarios, showing that:
(1) our inverse planning method robustly recognises the goals of non-ego vehicles;
(2) goal recognition leads to improved driving efficiency measured by driving time; and
(3) intuitive explanations for the predictions can be extracted to justify the system's decisions.
(Video showing IGP2 in action: \url{https://www.five.ai/igp2}.)

	\subsection{Scenarios}

\begin{figure}[t]
    \centering
    \includegraphics[width=0.2\textwidth]{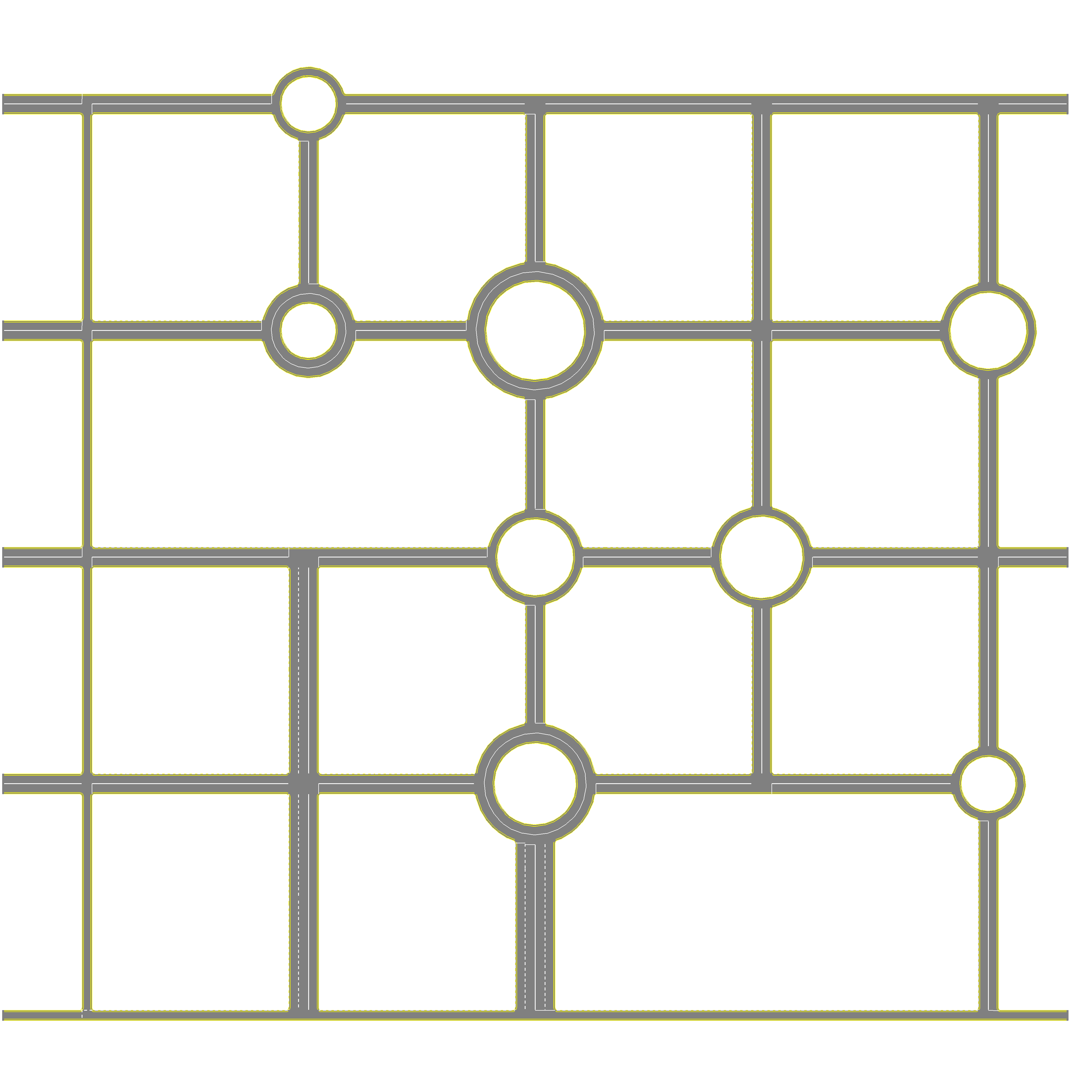}
    \hspace{10pt}
    \includegraphics[width=0.2\textwidth]{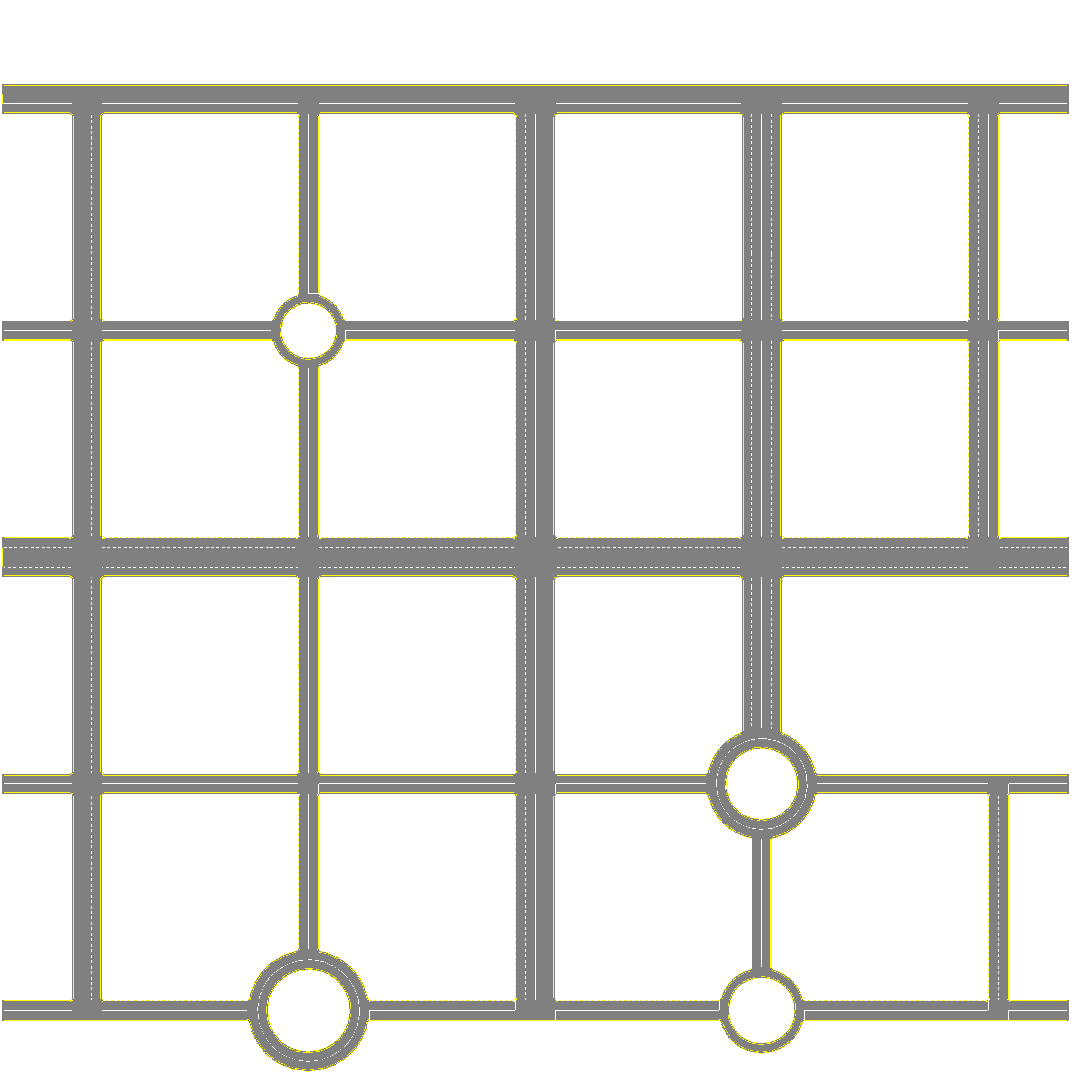}
    \caption{Town 1 and Town 2 layouts.}
    \label{fig:towns}
    \vspace{-1.0em}
\end{figure}

We use two sets of scenario instances. For in-depth analysis of goal recognition and planning, we use four defined local interaction scenarios shown in Figure~\ref{fig:scenarios}. For each of these scenarios, we generate 100 instances with randomly offset initial longitudinal positions ($\sim\hspace{-3pt}[-10,+10]$ meters) and initial speed sampled from range $[5,10]$ m/s for each vehicle including ego vehicle. Here the ego vehicle observes the whole scenario. To further assess IGP2's ability to complete full routes with random traffic, we use two random town layouts shown in Figure~\ref{fig:towns}. Each town spans an area of 0.16 square kilometers and consists of roads, crossings, and roundabouts with 2--4 lanes each. Each junction has one defined priority road. The ego vehicle's observation radius in towns is 50 meters. Non-ego vehicles are spawned within 25 meters outside the ego observation radius, with random road, lane, speed, and goal.
The total number of non-ego vehicles within the ego radius and spawning radius is kept at 8 to maintain a consistent medium-to-high level of traffic. In each town we generate 10 instances by choosing random routes for the ego vehicle to complete. The ego vehicle's goal is continually updated to be the outermost point on the route within the ego observation radius. In all simulations, the non-ego vehicles use manual heuristics to select from the maneuvers in Section~\ref{sec:maneuvers} to reach their goals. All vehicles use independent proportional controllers for acceleration and steering, and IDM \cite{idm} for automatic distance-keeping. Vehicle motion is simulated using a kinematic bicycle model.

        \subsection{Algorithms \& Parameters}
        \label{sec:exp-algos}

We compare the following algorithms in scenarios S1--S4. {\bf IGP2:} full system using goal recognition and MCTS. {\bf IGP2-MAP:} like IGP2, but MCTS uses only the most probable goal and trajectory for each vehicle. {\bf CVel:} MCTS without goal recognition, replaced by constant-velocity lane-following prediction after completion of current maneuver. {\bf CVel-Avg:} like CVel, but uses velocity averaged over the past 2 seconds. {\bf Cons:} like CVel, but using a conservative \man{give-way} maneuver which always waits until all oncoming vehicles on priority lanes have passed. In the town scenarios we focus on IGP2 and Cons, and additionally compare to {\bf SH-CVel} which works similarly to MPDM \cite{mpdm2017}: it simulates each macro action followed by a default \man{Continue} macro action, using CVel prediction for non-ego vehicles, then choosing the macro action with maximum estimated reward. (SH stands for ``short horizon'' as the search depth is effectively limited to 1.)

We simulate noisy maneuver detection (cf. Sec.~\ref{sec:mandetect}) by giving $0.9$ probability to the current executed maneuver of the non-ego vehicle and the rest uniformly to other maneuvers. Prior probabilities over non-ego goals are uniform. A* computes up to two predicted trajectories for each non-ego vehicle and goal. MCTS is run at a frequency of 1 Hz, performs $K=30$ simulations with a maximum search depth of $d_{max} = 5$, and uses $r_{coll} = r_{term} = -1$. We set $\lambda=10$ for velocity smoothing (cf. Eq.~\eqref{eq:smooth}).

        \subsection{Results}

            \subsubsection{Goal probabilities}

\begin{figure}[t]
    \centering
    \includegraphics[height=0.175\textheight]{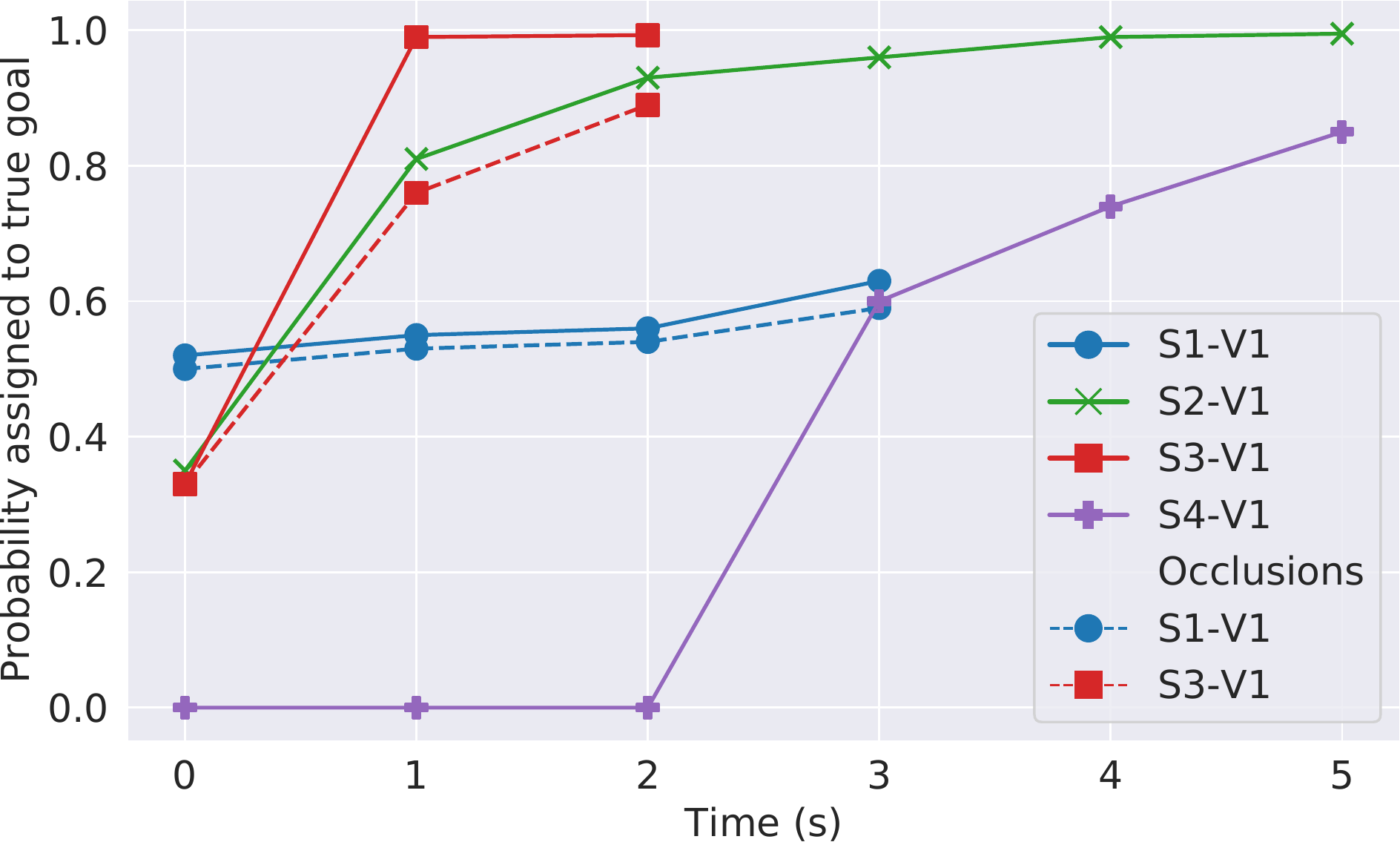}
    \caption{Average probability given to true goal of selected vehicles in scenarios S1--S4. Note: lines for S1/S3 are shorter than indicated in Tab.~\ref{tab:times-scenarios} since possible vehicle goals change after exit points are reached and we only show lines for initial possible goals.}
    \label{fig:goalprobs}
    \vspace{-0.9em}
\end{figure}

Figure~\ref{fig:goalprobs} shows the average probability over time assigned to the true goal in scenarios S1--S4. In all tested scenario instances we observe that the probability increases with growing evidence and at different rates depending on random scenario initialisation. Snapshots of goal probabilities (shown as bar plots) associated with the non-ego's most probable current maneuver can be seen in Figure~\ref{fig:scenarios}. We also tested the method's robustness to missing segments in the observed trajectory of a vehicle. In scenarios S1 and S3 we removed the entire \man{lane-change} maneuver from the observed trajectory (but keeping the short lane-follow segment before the lane change). To deal with occlusion, we applied A* search before the beginning of each missing segment to reach the beginning of the next observed segment, thereby ``filling the gaps'' in the trajectory. Afterwards we applied velocity smoothing to the reconstructed trajectory. The results are shown as dashed lines in Figure~\ref{fig:goalprobs}, showing that even under significant occlusion the method is able to correctly recognise a vehicle's goal.

            \subsubsection{Driving times}

Table~\ref{tab:times-scenarios} shows the average driving times required of each algorithm in scenarios S1--S4. Goal recognition enabled IGP2 and IGP2-MAP to reduce their driving times. {\bf (S1)} All algorithms change lanes to avoid being slowed down by V1, leading to same driving times, however IGP2 and IGP2-MAP initiate the lane change before all other algorithms by recognising V1's intended goal. {\bf (S2)} Cons waits for $V_1$ to clear the lane, which in turn must wait for $V_2$ to pass. IGP2 and IGP2-MAP anticipate this behaviour, allowing them to enter the road earlier. CVel and CVel-Avg wait for $V_1$ to reach near-zero velocity. {\bf (S3)} IGP2 and IGP2-MAP are able to enter early as they recognise $V_1$'s goal to exit the roundabout, while CVel, CVel-Avg, and Cons wait for $V_1$ to exit. {\bf (S4)} Cons waits until $V_1$ decides to close the gap after which the ego can enter the road. IGP2 and IGP2-MAP recognise $V_1$'s goal and merge in front.

IGP2-MAP achieved shorter driving times than IGP2 on some scenario instances (such as S3 and S4). This is because IGP2-MAP commits to the most-likely goal and trajectory of other vehicles, while IGP2 also considers residual uncertainty about goals and trajectories which may lead MCTS to select more cautious actions in some situations. The limitation of IGP2-MAP can be seen when simulating unexpected (irrational) behaviours in other vehicles. To test this, we compared IGP2 and IGP2-MAP on instances from S3 and S4 which were modified such that V1, after slowing down, suddenly accelerates and continues straight (rather than exiting as in S3, or stopping as in S4). In these cases we observed a 2-3\% collision rate for IGP2-MAP (in all collisions, V1 collided into the ego) while IGP2 produced no collisions. These results show that IGP2 exhibits safer driving than IGP2-MAP by accounting for uncertainty over goals and trajectories.

\begin{figure}[t]
    \centering
    \includegraphics[height=0.16\textheight]{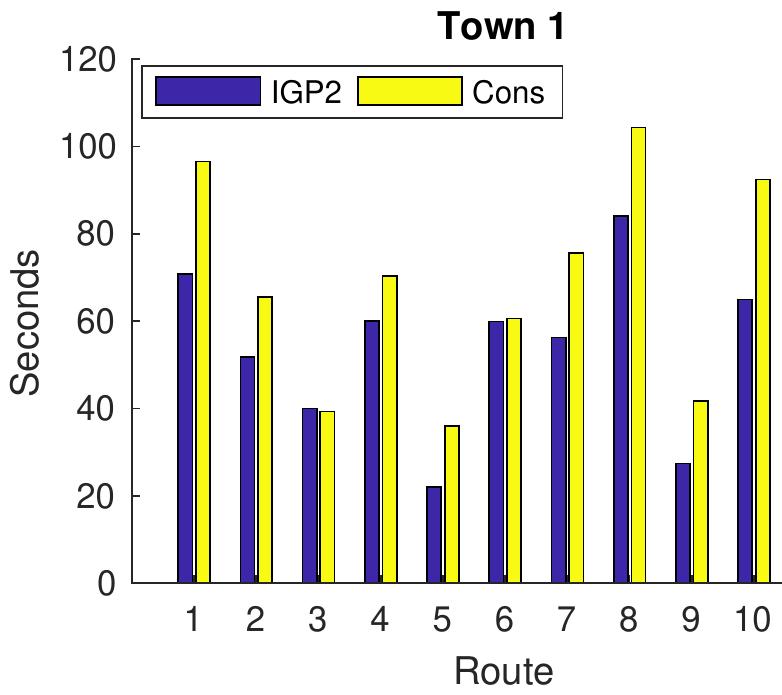}
    \hfill
    \includegraphics[height=0.16\textheight]{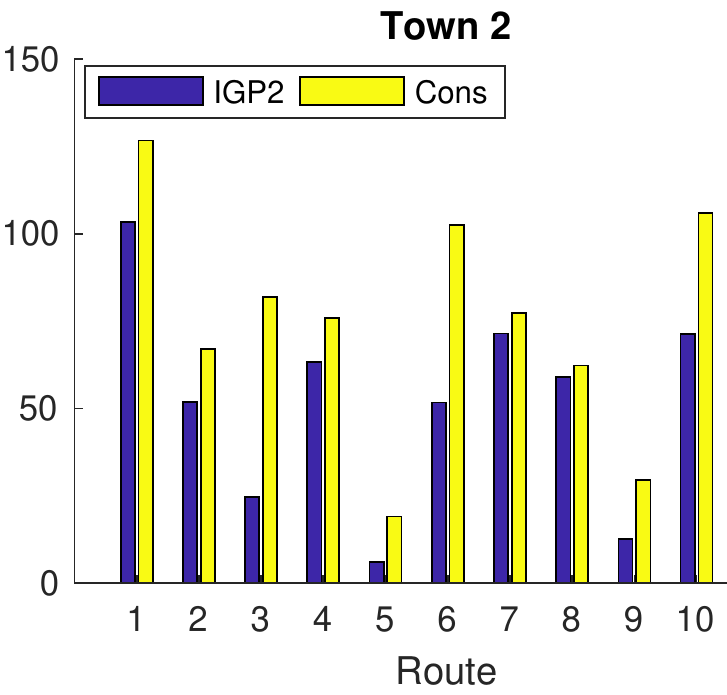}
    \caption{Driving times (seconds) of IGP2 and Cons for 10 routes in Town 1 and Town 2.}
    \label{fig:times-towns}
    \vspace{-0.9em}
\end{figure}

Figure~\ref{fig:times-towns} shows the driving times of IGP2 and Cons for the routes in the two towns. Both algorithms completed all of the routes. Goal recognition allowed IGP2 to reduce its driving times substantially by exploiting multiple opportunities for proactive lane changes and road/junction entries. In contrast, Cons exhibited more conservative driving and often waited considerably longer at junctions or before taking a turn until traffic cleared up. SH-CVel was unable to complete any of the given routes, as its short planning horizon often caused it to take a wrong turn (thus failing the instance).

            \subsubsection{Interpretability}

We are able to extract intuitive explanations for the predictions and decisions made by IGP2. The explanations are given in the caption of Figure~\ref{fig:scenarios}.

\begin{table}[t]
	\small
	\center
	\begin{tabular}{| @{\ }l@{\ } | @{\ }c@{\ } | @{\ }c@{\ } | @{\ }c@{\ } | @{\ }c@{\ } |}
		\hline
					    & S1				& S2				& S3				& S4 \\
		\hline
		IGP2			& $5.97 \pm .02$	& $7.24 \pm .05$	& $8.54 \pm .05$	& $10.83 \pm .03$ \\
		IGP2-MAP	    & $5.99 \pm .02$	& $7.23 \pm .05$	& $8.36 \pm .06$	& $10.40 \pm .03$ \\
		CVel			& $6.04 \pm .03$	& $9.80 \pm .17$	& $10.49 \pm .09$	& $12.83 \pm .03$ \\
		CVel-Avg		& $6.01 \pm .02$	& $11.31 \pm .17$	& $10.49 \pm .09$	& $13.59 \pm .02$ \\
		Cons			& $6.01 \pm .02$	& $12.89 \pm .03$	& $10.90 \pm .04$	& $16.78 \pm .02$ \\
		\hline
	\end{tabular}
	\caption{Average driving time (seconds) required to complete scenario instances from S1--S4, with standard error.}
	\label{tab:times-scenarios}
\end{table}

    \section{Conclusion}
    \label{sec:conc}

We proposed an autonomous driving system, IGP2, which integrates planning and prediction over extended horizons by reasoning about the goals of other vehicles via rational inverse planning. Evaluation in diverse urban driving scenarios showed that IGP2 robustly recognises the goals of non-ego vehicles, resulting in improved driving efficiency while allowing for intuitive interpretations of the predictions to explain the system's decisions. IGP2 is general in that it uses relatively standard planning techniques that could be replaced with other techniques (e.g. POMDP-based planners \cite{despot2013}), and the general principles underlying our approach could be
applied to other domains in which mobile robots interact with other robots/humans. Important future directions include goal recognition in the presence of occluded objects which can be seen by the non-ego vehicle but not the ego vehicle, and accounting for human irrational biases \cite{sadigh2020,hu2019generic}.

    \bibliographystyle{IEEEtran}
    \bibliography{icra2021}

\end{document}